\pdfoutput=1
\documentclass[11pt]{article}
\usepackage{nodalida2019}
\usepackage{times}
\usepackage{url}
\usepackage{latexsym}

\usepackage{hyperref}

\usepackage{numprint}
\usepackage{amsmath}
\usepackage{amssymb}
\usepackage{comment}
\usepackage{booktabs}
\usepackage{subfig}
\usepackage{array,graphicx}
\usepackage{booktabs}
\usepackage{pifont}
\usepackage{tikz}

\aclfinalcopy 

\newcommand{\Cross}{$\mathbin{\tikz [x=1.4ex,y=1.4ex,line width=.2ex, red] \draw (0,0) -- (1,1) (0,1) -- (1,0);}$}%

\title{Mark my Word: \\ A Sequence-to-Sequence Approach to Definition Modeling}

\author{Timothee Mickus \\
Universit\'e de Lorraine \\
 CNRS, ATILF \\ 
 {\tt tmickus@atilf.fr} \\\And
 Denis Paperno \\
 Utrecht University \\
 {\tt d.paperno@uu.nl} 
  \\\And
 Mathieu Constant \\
 Universit\'e de Lorraine \\
 CNRS, ATILF \\ 
 {\tt mconstant@atilf.fr} \\}

\date{}

\begin{document}
\maketitle

\begin{abstract}Defining words in a textual context is a useful task both for practical purposes and for gaining insight into distributed word representations.\ 
Building on the distributional hypothesis, we argue here that 
the most natural formalization of definition modeling is to treat it as a sequence-to-sequence task, rather than a word-to-sequence task: given an input sequence with a highlighted word, generate a contextually appropriate definition for it. We implement this approach in a Transformer-based sequence-to-sequence model. 
Our proposal allows to train contextualization and definition generation in an end-to-end fashion, which is a conceptual improvement over earlier works. We achieve state-of-the-art results both in contextual and non-contextual definition modeling.
\end{abstract}


\section{Introduction} \label{sec:intro}

The task of \textsl{definition modeling}, introduced by \citet{Noraset2017DefinitionML}, consists in generating the dictionary definition of a specific word: for instance, given the word ``\textit{monotreme}'' as input, the system would need to produce a definition such as ``\textit{any of an order (Monotremata) of egg-laying mammals comprising the platypuses and echidnas}''.\footnote{Definition from Merriam-Webster.} Following the tradition set by lexicographers, we call the word being defined a \textit{definiendum} (pl.\ \textit{definienda}), whereas a word occurring in its definition is called a \textit{definiens} (pl.\ \textit{definientia}).



Definition modeling can prove useful in a variety of applications. Systems trained for the task may generate dictionaries for low resource languages, or extend the coverage of existing lexicographic resources where needed, e.g.\ of domain-specific vocabulary. 
Such systems may also be able to provide reading help 
by giving definitions for words in the text.

A major intended application of definition modeling is the explication and evaluation of distributed lexical representations, also known as word embeddings \citep{Noraset2017DefinitionML}. This evaluation procedure is based on the postulate that the meaning of a word, as is captured by its embedding, should be convertible into a human-readable dictionary definition. How well the meaning is captured must impact the ability of the model to reproduce the definition, and therefore embedding architectures can be compared according to their downstream performance on definition modeling. This intended usage 
motivates the requirement that definition modeling architectures take as input the embedding of the \textit{definiendum} and not retrain it.

From a theoretical point of view, usage of word embeddings as representations of meaning \citep[cf.][for an overview]{lenci2018distributional,Boleda2019DSandLT} is motivated by the distributional hypothesis \citep{Harris54}. This framework holds that meaning 
can be inferred from the linguistic context of the word, usually seen as co-occurrence data. The context of usage is even more crucial for characterizing meanings of ambiguous or polysemous words: a definition that does not take disambiguating context into account will be of limited use \citep{Gadetsky18WordDefGen}.

We argue that definition modeling should preserve the link between the \textit{definiendum} and its context of occurrence. The most natural approach to this task is to treat it as a \textsl{sequence-to-sequence} task, rather than a \textsl{word-to-sequence} task: given an input sequence with a highlighted word, generate a contextually appropriate definition for it (cf.\ sections \ref{sec:seq2seq} \& \ref{sec:formal}). We implement this approach in a Transformer-based sequence-to-sequence model that achieves state-of-the-art performances (sections \ref{sec:eval} \& \ref{sec:quali}).

\section{Related Work} \label{sec:relw}
In their seminal work on definition modeling, \citet{Noraset2017DefinitionML} likened systems generating definitions to language models, which can naturally be used to generate arbitrary text. They built a sequential \textsc{lstm} seeded with the embedding of the \textit{definiendum}; its output at each time-step was mixed through a gating mechanism with a feature vector derived from the \textit{definiendum}. 

\citet{Gadetsky18WordDefGen} stressed that a \textit{definiendum} outside of its specific usage context is ambiguous between all of its possible definitions.
They proposed to first compute the AdaGram vector \citep{Bartunov15Adagram} for the \textit{definiendum}, to then disambiguate it using a gating mechanism learned over contextual information, and finally to run a language model over the sequence of \textit{definientia} embeddings prepended with the disambiguated \textit{definiendum} embedding. 

In an attempt to produce a more interpretable model, \citet{Chang18xSense} 
map the \textit{definiendum} to a sparse vector representation. 
Their architecture comprises four modules. The first encodes the context in a sentence embedding, the second converts the \textit{definiendum} into a sparse vector, the third combines the context embedding and the sparse representation, passing them on to 
the last module  
which generates the definition.

Related to these works, \citet{Yang2019Sememes4CDM} specifically tackle definition modeling in the context of Chinese---whereas all previous works on definition modeling studied English. In a Transformer-based architecture, they incorporate ``sememes'' as part of the representation of the \textit{definiendum} to generate definitions. 

On a more abstract level, definition modeling is related to research on the analysis and evaluation of word embeddings \citep[][e.g.]{Levy2014a,Levy2014,Arora16LinearWordsenses,batchkarov2016critiquewordsim,swinger2018biases}.
It also relates to other works associating definitions and embeddings, 
like the ``reverse dictionary task'' \citep{Hill16DictRep}---retrieving the \textit{definiendum} knowing its definition, 
which can be argued to be the opposite of definition modeling---
or works that derive embeddings from definitions
\citep{Wang15LexEmbLexKno,Tissier2017Dict2vecL,Bosc18AutoencodeDefs}. 

\section{Definition modeling as a sequence-to-sequence task} \label{sec:seq2seq}

\citet{Gadetsky18WordDefGen} remarked that words are often ambiguous or polysemous, and thus generating a correct definition requires that we either use sense-level representations, or that we disambiguate the word embedding of the \textit{definiendum}.
The disambiguation that \citet{Gadetsky18WordDefGen} proposed was based on a contextual cue---ie. a short text fragment. As \citet{Chang18xSense} notes, the cues in \citeauthor{Gadetsky18WordDefGen}'s \citeyearpar{Gadetsky18WordDefGen} dataset do not necessarily contain the \textit{definiendum} or even an inflected variant thereof. For instance, one training example disambiguated the word ``\textit{fool}'' using the cue ``\textit{enough horsing around---let's get back to work!}''. 

Though the remark that \textit{definienda} must be disambiguated is pertinent, the more natural formulation of such a setup would be to disambiguate the \textit{definiendum} using its actual context of occurrence. In that respect, the \textit{definiendum} and the contextual cue would form a linguistically coherent sequence, and thus it would make sense to encode the context together with the \textit{definiendum}, rather than to merely rectify the \textit{definiendum} embedding using a contextual cue. Therefore, definition modeling is by its nature a sequence-to-sequence task: mapping contexts of occurrence of \textit{definienda} to definitions.


This remark can be linked to the distributional hypothesis \citep{Harris54}. 
The distributional hypothesis suggests that a word's meaning can be inferred from its context of usage; or, more succinctly, that ``you shall know a word by the company it keeps'' \citep{firth1957}. When applied to definition modeling, the hypothesis can be rephrased as follows: the correct definition of a word can only be given  when knowing in what linguistic context(s) it occurs. %
Though different kinds of linguistic contexts have been suggested throughout the literature, we remark here that sentential context may sometimes suffice to guess the meaning of a word that we don't know \citep{Lazaridou2017MultimodalWM}. 
Quoting from the example above, the context ``\textit{enough \underline{\hspace{1cm}} around---let's get back to work!}'' sufficiently characterizes the meaning of the omitted verb to allow for an approximate definition for it even if the blank is not filled \citep{Taylor53Cloze,Devlin18Bert}.


This reformulation can appear contrary to the original proposal by \citet{Noraset2017DefinitionML}, which conceived definition modeling as a ``word-to-sequence task''. They argued for an approach related to, though distinct from sequence-to-sequence architectures. Concretely, a specific encoding procedure was applied to the \textit{definiendum}, so that it could be used as a feature vector during generation. In the simplest case, vector encoding of the \textit{definiendum} consists in looking up its vector in a vocabulary embedding matrix. 

We argue that the whole context of a word's usage should be accessible to the generation algorithm rather than a single vector.
To take a more specific case of verb definitions, we observe that context explicitly represents argument structure, which is obviously useful when defining the verb. There is no guarantee that a single embedding, even if it be contextualized, would preserve this wealth of information---that is to say, that you can cram all the information pertaining to the syntactic context into a single vector.

Despite some key differences, all of the previously proposed architectures we are aware of \citep{Noraset2017DefinitionML,Gadetsky18WordDefGen,Chang18xSense,Yang2019Sememes4CDM} followed a pattern similar to sequence-to-sequence models. They all implicitly or explicitly used distinct submodules to encode the \textit{definiendum} and to generate the \textit{definientia}.
In the case of \citet{Noraset2017DefinitionML}, the encoding was the concatenation of the embedding of the \textit{definiendum}, a vector representation of its sequence of characters derived from a character-level \textsc{cnn}, and its ``hypernym embedding''. \citet{Gadetsky18WordDefGen} used a sigmoid-based gating module to tweak the \textit{definiendum} embedding. The architecture proposed by \citet{Chang18xSense} is comprised of four modules, only one of which is used as a decoder: the remaining three are meant to convert the \textit{definiendum} as a sparse embedding, select some of the sparse components of its meaning based on a provided context, and encode it into a representation adequate for the decoder.

Aside from theoretical implications, there is another clear gain in considering definition modeling as a sequence-to-sequence task. Recent advances in embedding designs have introduced contextual embeddings \citep{McCann17CoVe,Peters18ELMo,Devlin18Bert}; and these share the particularity that they are a ``function of the entire sentence'' \citep{Peters18ELMo}: in other words, vector representations are assigned to tokens rather than to word types, and moreover semantic information about a token can be distributed over other token representations. 
%
%
To extend definition modeling to contextual embeddings therefore requires that we devise architectures able to encode a word in its context; in that respect sequence-to-sequence architectures are a natural choice.

A related point is that not all \textit{definienda} are comprised of a single word: multi-word expressions include multiple tokens, yet receive a single definition. Word embedding architectures generally require a pre-processing step to detect these expressions and merge them into a single token. However, as they come with varying degrees of semantic opacity \citep{cordeiro2016MWECompEmbeddingsHardTime}, a definition modeling system would benefit from directly accessing the tokens they are made up from. Therefore, if we are to address the entirety of the language and the entirety of existing embedding architectures in future studies, reformulating definition modeling as a sequence-to-sequence task becomes a necessity.


\section{Formalization} \label{sec:formal}

A sequence-to-sequence formulation of definition modeling can formally be seen as a mapping between contexts of occurrence of \textit{definienda} and their corresponding definitions. It moreover requires that the \textit{definiendum} be formally distinguished from the remaining context: otherwise the definition could not be linked to any particular word of the contextual sequence, and thus would need to be equally valid for any word of the contextual sequence.

We formalize definition modeling as mapping to sequences of \textit{definientia} from sequences of pairs $\langle w_1,i_1\rangle,~ \dots,~ \langle w_n,i_n\rangle$ , where $w_k$ is the $k$\textsuperscript{th} word in the input and $i_k\in\{0,1\}$ indicates whether the $k$\textsuperscript{th} token is to be defined. As only one element of the sequence should be highlighted, we expect the set of all indicators to contain only two elements: the one, $i_d=1$, to mark the \textit{definiendum}, the other, $i_c=0$, to mark the context; this entails that we encode this marking using one bit only.\footnote{Multiple instances of the same \textit{definiendum} within a single context should all share a single definition, and therefore could theoretically all be marked using the \textit{definiendum} indicator $i_d=1$. Likewise the words that make up a multi-word expression should all be marked with this $i_d$ indicator. In this work, however, we only mark a single item; in cases when multiple occurrences of the same \textit{definiendum} were attested, we simply marked the first occurrence.}

To treat definition modeling as a sequence-to-sequence task, the information from each pair $\langle w_k,i_k\rangle$ has to be integrated into a single representation $\vec{\textit{marked}_k}$:
\begin{align} \label{eq:general}
    \vec{\textit{marked}_k} = \textit{mark}(i_k,\vec{w_k})
\end{align}
This marking function can theoretically take any form. Considering that definition modeling uses the embedding of the definiendum $\vec{w_d} = e(w_d)$, in this work we study a multiplicative and an additive mechanism, as they are conceptually the simplest form this marking can take in a vector space. They are given schematically in Figure~\ref{fig:marking-schemes}, and formally defined as:
\begin{align}
    \vec{\textit{marked}_k^{\times}} &= i_k \times \vec{w_k} \\
    \vec{\textit{marked}_k^{+}} &= \textit{e}(i_k) + \vec{w_k}
\end{align}

The last point to take into account is where to set the marking. Two natural choices are to set it either before or after encoded representations were obtained. We can formalize this using either of the following equation, with $\mathcal{E}$ the model's encoder:
\begin{align} \label{eq:pre-or-post}
    \vec{\textit{marked}_k^\text{~ after}} & = \textit{mark}(i_k, \mathcal{E}(\vec{w_k})) \nonumber \\ 
    \vec{\textit{marked}_k^\text{~ before}} &= \mathcal{E}(\textit{mark}(i_k, \vec{w_k}))
\end{align}

\subsection{Multiplicative marking: \textsc{Select}} \label{sec:slct-sub}

The first option we consider is to use scalar multiplication to distinguish the word to define. In such a scenario, the marked token encoding is
\begin{equation}
    \setcounter{equation}{2}
    \vec{\textit{marked}_k^{\times}} = i_k \times \vec{w_k}
\end{equation}
As we use bit information as indicators, this form of marking entails that only the representation of the \textit{definiendum} be preserved and that all other contextual representations are set to $\vec{0} = (0,~ \cdots,~ 0)$: thus multiplicative marking amounts to selecting just the \textit{definiendum} embedding and discarding other token embeddings. The contextualized \textit{definiendum} encoding bears the trace of its context, but detailed information is irreparably lost.  
Hence, we refer to such an integration mechanism as a \textsc{Select} marking of the \textit{definiendum}. 

When to apply marking, as introduced by eq. \ref{eq:pre-or-post}, is crucial when using the multiplicative marking scheme \textsc{Select}. Should we mark the \textit{definiendum} before encoding, 
then only the \textit{definiendum} embedding is passed into the encoder: the resulting system provides out-of-context definitions, like in \citet{Noraset2017DefinitionML} where the definition is not linked to the context of a word but to its \textit{definiendum} only. 
For context to be taken into account under the multiplicative strategy, tokens $w_k$ must be encoded and contextualized before integration with the indicator $i_k$.

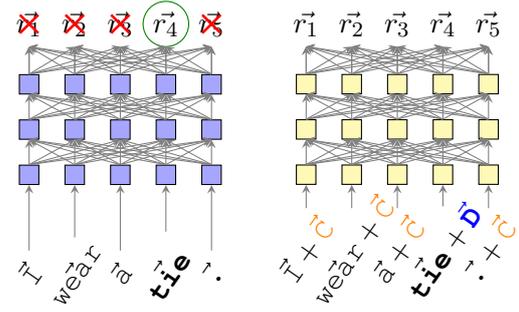
\begin{figure}[t]
    \centering
    \subfloat[
    \label{fig:subtractive-selection-encoder} \textsc{Select}: Selecting from encoded items. Items are contextualized and the \textit{definiendum} is singled out from them.]{
    \begin{tikzpicture}[
 decstyle/.style={rectangle,draw,fill=white!65!blue}, 
 scale=1]
 \foreach \x in {0,...,4}
 \foreach \y in {0,...,2} 
 {\node [decstyle] (d\x\y) at (.6*\x, .6*\y) {};} 
\foreach \t [count=\ti from 0] in {{$\vec{r_1}$}, {$\vec{r_2}$}, {$\vec{r_3}$}, {$\vec{r_4}$}, {$\vec{r_5}$}}
 {
 \node[] (d\ti3) at (.6*\ti,.6 *3 + 0.2) {\t};
 }
 \foreach \x in {0,...,4} \draw[-stealth,gray] (d\x2.north)--(d\x3.south);
 
 \foreach \x in {0,...,2} \node at (.6*\x,.6 * 3 + 0.2) {\Cross};
 \node at (.6*4,.6 * 3 + 0.2) {\Cross};
 \node[draw, circle, minimum size=.6cm, green!50!black] at (.6*3,.6 * 3 + 0.2) {};

 \foreach \y [count=\yi] in {0,...,2}
 \foreach \x in {0,...,4}
 \foreach \xi in {0,...,4}
 \draw[-stealth,gray] (d\x\y.north)--(d\xi\yi.south);

\foreach \t [count=\ti from 0] in {{$\vec{\texttt{I}}$}, {$\vec{\texttt{wear}}$}, {$\vec{\texttt{a}}$}, {$\vec{\textbf{\texttt{tie}}}$}, {$\vec{\textbf{\texttt{.}}}$}}
 {
 \node[rotate=55] (e\ti0) at (.6*\ti,-1.31) {\t};
 }
 \foreach \x in {0,...,4} \draw[-stealth,gray] (e\x0)--(d\x0.south);

\end{tikzpicture}} ~ 
\subfloat[\label{fig:additive-selection-encoder} \textsc{Add}: Additive marking in encoder. Context items and \textit{definiendum} are marked by adding dedicated embeddings.]{
\begin{tikzpicture}[
 decstyle/.style={rectangle,draw,fill=white!65!yellow}, 
 scale=1]

 \foreach \x in {0,...,4}
 \foreach \y in {0,...,2} 
 {\node [decstyle] (d\x\y) at (.6*\x, .6*\y) {};} 
\foreach \t [count=\ti from 0] in {{$\vec{r_1}$}, {$\vec{r_2}$}, {$\vec{r_3}$}, {$\vec{r_4}$}, {$\vec{r_5}$}}
 {
 \node[] (d\ti4) at (.6*\ti,.6 * 3 + 0.2) {\t};
 }
 \foreach \x in {0,...,4} \draw[-stealth,gray] (d\x2.north)--(d\x3.south);
 \foreach \y [count=\yi] in {0,...,2}
 \foreach \x in {0,...,4}
 \foreach \xi in {0,...,4}
 \draw[-stealth,gray] (d\x\y.north)--(d\xi\yi.south);

\foreach \t [count=\ti from 0] in {{$\vec{\texttt{I}}+\textcolor{orange}{\vec{\texttt{C}}}$}, {$\vec{\texttt{wear}}+\textcolor{orange}{\vec{\texttt{C}}}$}, {$\vec{\texttt{a}}+\textcolor{orange}{\vec{\texttt{C}}}$}, {$\vec{\textbf{\texttt{tie}}}+\textcolor{blue}{\vec{\textbf{\texttt{D}}}}$}, {$\vec{\texttt{.}}+\textcolor{orange}{\vec{\texttt{C}}}$}}
 {
 \node[rotate=55] (e\ti0) at (.6*\ti,-1) {\t};
 }
 \foreach \x in {0,...,4} \draw[-stealth,gray] (e\x0)--(d\x0.south);

\end{tikzpicture}}
    \caption{Additive vs. multiplicative integration}
    \label{fig:marking-schemes}
\end{figure}

Figure~\ref{fig:subtractive-selection-encoder} presents the contextual \textsc{Select} mechanism visually.
It consists in coercing the decoder to attend only to the contextualized representation for the \textit{definiendum}. To do so, we encode the full context and then select only the encoded representation of the \textit{definiendum}, dropping the rest of the context, before running the decoder. In the case of the Transformer architecture, this is equivalent to using a multiplicative marking on the encoded representations: vectors that have been zeroed out are ignored during attention and thus cannot influence the behavior of the decoder.

This \textsc{Select} approach 
may seem intuitive and naturally interpretable, as it directly controls what information is passed to the decoder---we carefully select only the contextualized \textit{definiendum}, thus the only remaining zone of uncertainty would be how exactly contextualization is performed. It also seems to provide a strong and reasonable bias for training the definition generation system. 
Such an approach, however, is not guaranteed to excel: forcibly omitted context could contain important information that might not be easily incorporated in the \textit{definiendum} embedding.

Being simple and natural, the \textsc{Select} approach resembles architectures like that of \citet{Gadetsky18WordDefGen} and \citet{Chang18xSense}: the full encoder is dedicated to altering the embedding of the \textit{definiendum} on the basis of its context; in that, the encoder may be seen as a dedicated contextualization sub-module. 

\subsection{Additive marking: \textsc{Add}} \label{sec:add-sub}

We also study an additive mechanism shown in Figure~\ref{fig:additive-selection-encoder} (henceforth \textsc{Add}). It concretely consists in embedding the word $w_k$ and its indicator bit $i_k$ in the same vector space and adding the corresponding vectors:
\begin{equation}
    \setcounter{equation}{3}
    \vec{\textit{marked}_k^{+}} = \textit{e}(i_k) + \vec{w_k}
\end{equation}
%
%
In other words, under \textsc{Add} we distinguish the \textit{definiendum} by adding a vector $\vec{\texttt{D}}$ 
to the \textit{definiendum} embedding, and another vector $\vec{\texttt{C}}$ 
to the remaining context token embeddings; both markers $\vec{\texttt{D}}$ and $\vec{\texttt{C}}$  are learned during training. In our implementation, markers are added to the input of the encoder, so that the encoder has access to this information; 
we leave the question of whether to integrate indicators and words at other points of the encoding process, as suggested in eq. \ref{eq:pre-or-post}, to future work.

Additive marking of substantive features has its precedents. For example, \textsc{Bert} embeddings \citep{Devlin18Bert} are trained using two sentences at once as input; sentences are distinguished with added markers called ``segment encodings''. 
Tokens from the first sentence are all marked with an added vector $\vec{\texttt{seg}_A}$, whereas tokens from second sentences are all marked with an added vector $\vec{\texttt{seg}_B}$. The main difference here is that we only mark one item with the marker $\vec{D}$, while all others are marked with $\vec{C}$.

This \textsc{Add} marking is more expressive than the \textsc{Select} architecture. Sequence-to-sequence decoders typically employ an attention to the input source \citep{Bahdanau14Attention}, which corresponds to a re-weighting of the encoded input sequence based on a similarity between the current state of the decoder (the `query') and each member of the input sequence (the `keys'). This re-weighting is normalized with a softmax function, producing a probability distribution over keys. 
However, both non-contextual definition modeling and the  \textsc{Select} approach produce singleton encoded sequences: in such scenarios the attention mechanism assigns a single weight of 1 and thus devolves into a simple linear transformation of the value and makes the attention mechanism useless. 
Using an additive marker, rather than a selective mechanism, will prevent this behavior.

\section{Evaluation} \label{sec:eval}

We implement several sequence to sequence models with the Transformer architecture \citep{Vaswani17}, building on the Open\textsc{nmt} library \citep{opennmt2017} with adaptations and modifications when necessary.\footnote{Code \& data are available at the following \textsc{url}:  \href{https://github.com/TimotheeMickus/onmt-selectrans}{https://github.com/TimotheeMickus/onmt-selectrans}.
} 
Throughout this work, we use GloVe vectors \citep{Pennington2014} and freeze weights of all embeddings 
for a fairer comparison with previous models; words not in GloVe but observed in train or validation data and missing \textit{definienda} in our test sets were randomly initialized with components drawn from a normal distribution $\mathcal{N}(0, 1)$.

We train a distinct model for each dataset.
We batch examples by 8,192, using gradient accumulation to circumvent \textsc{gpu} limitations.
We optimize the network using Adam with $\beta_1 = 0.99$, $\beta_2 = 0.998$, a learning rate of 2, label smoothing of 0.1, Noam exponential decay with 2000 warmup steps, and dropout rate of 0.4.
The parameters are initialized using Xavier.
Models were trained for up to 120,000 steps with checkpoints at each 1000 steps; we stopped training if perplexity on the validation dataset stopped improving. We 
report results from checkpoints performing best on validation. 

\subsection{Implementation of the Non-contextual Definition Modeling System}
In non-contextual definition modeling, \textit{definienda} are mapped directly to definitions. 
As the source corresponds only to the \textit{definiendum}, we conjecture that few parameters are required for the encoder. We use 1 layer for the encoder, 6 for the decoder, 300 dimensions per hidden representations and 6 heads for multi-head attention. 
We do not share vocabularies between the encoder and the decoder: therefore output tokens can only correspond to words attested as \textit{definientia}.\footnote{In our case, not sharing vocabularies prevents the model from considering rare words only used as \textit{definienda}, such as ``\textit{penumbra}'' as potential outputs, and was found to improve performances.}


The dropout rate and warmup steps number were set using a hyperparameter search on the dataset from \citet{Noraset2017DefinitionML}, during which encoder and decoder vocabulary were merged for computational simplicity and models stopped after 12,000 steps. We first fixed dropout to 0.1 and tested warmup step values between 1000 and 10,000 by increments of 1000, then focused on the most promising span (1000--4000 steps) and exhaustively tested dropout rates from 0.2 to 0.8 by increments of 0.1.


\subsection{Implementation of Contextualized Definition Modeling Systems}
To compare the effects of the two integration strategies that we discussed in section \ref{sec:formal}, we implement both the additive marking approach (\textsc{Add}, cf. section \ref{sec:add-sub}) and the alternative `encode and select' approach (\textsc{Select}, cf. section \ref{sec:slct-sub}).
To match with the complex input source, we define encoders with 6 layers; we reemploy the set of hyperparameters previously found for the non-contextual system. Other implementation details, initialization strategies and optimization algorithms are kept the same as described above for the non-contextual version of the model.

We stress that the two approaches we compare for contextualizing the {definiendum} are applicable to almost any sequence-to-sequence neural architecture with an attention mechanism to the input source.\footnote{For best results, the \textsc{Select} mechanism should require a bi-directional encoding mechanism.} Here we chose to rely on a Transformer-based architecture \citep{Vaswani17}, which has set the state of the art in a wide range of tasks, from language modeling \citep{dai19tfxl} to machine translation \citep{ott18scalingNMT}. It is therefore expected that the Transformer architecture will also improve performances for definition modeling, if our arguments for treating it as a sequence to sequence task are on the right track.

\subsection{Datasets} \label{sec:data}

We train our models on three distinct datasets, which are all borrowed or adapted from previous works on definition modeling. As a consequence, our experiments focus on the English language.
The dataset of \citet{Noraset2017DefinitionML} (henceforth \textbf{$D_\text{Nor}$}) maps \textit{definienda} to their respective \textit{definientia}, as well as additional information not used here. 
In the dataset of  \citet{Gadetsky18WordDefGen} (henceforth \textbf{$D_\text{Gad}$}),  each example consists of a \textit{definiendum}, the \textit{definientia} for one of its meanings and a contextual cue sentence. 
$D_\text{Nor}$ contains on average shorter definitions than $D_\text{Gad}$. Definitions in $D_\text{Nor}$ have a mean length 
of $6.6$ and a standard deviation of $5.78$, whereas those in $D_\text{Gad}$ have a mean length of $11.01$ and a standard deviation of $6.96$. 

\citet{Chang18xSense} stress that the dataset $D_\text{Gad}$ includes many examples where the \textit{definiendum} is absent from the associated cue. About half of these cues doe not contain an exact match for the corresponding \textit{definiendum}, but up to 80\% contains either an exact match or an inflected form of the \textit{definiendum} according to lemmatization by the \textsc{nltk} toolkit \citep{nltk}. 
To cope with this problematic characteristic, we converted the dataset into the word-in-context format assumed by our model by concatenating the \textit{definiendum} with the cue. To illustrate this, consider the actual input from $D_\text{Gad}$ comprised of the \textit{definiendum} ``\textit{fool}'' and its associated cue ``\textit{enough horsing around---let's get back to work!}'': to convert this into a single sequence, we simply prepend the \textit{definiendum} to the cue, which results in the sequence ``\textit{fool enough horsing around---let's get back to work!}''.
Hence the input sequences of $D_{Gad}$ do not constitute linguistically coherent sequences, but it does guarantee that our sequence-to-sequence variants have access to the same input as previous models; therefore the inclusion of this dataset in our experiments is intended mainly for comparison with previous architectures. We also note that this conversion procedure entails that our examples have a very regular structure: the word marked as a \textit{definiendum} is always the first word in the input sequence.

Our second strategy was to restrict the dataset by selecting only cues where the \textit{definiendum} (or its inflected form) is present. 
The curated dataset (henceforth \textbf{$D_\text{Ctx}$}) contains 78,717 training examples, 9,413 for validation and 9,812 for testing. 
In each example, the first occurrence of the \textit{definiendum} is annotated as such.
$D_\text{Ctx}$ thus differs from $D_\text{Gad}$ in two ways: some definitions have been removed, and the exact citation forms of the \textit{definienda} are not given.  Models trained on $D_\text{Ctx}$ implicitly need to lemmatize the \textit{definiendum}, since inflected variants of a given word are to be aligned to a common representation; thus they are not directly comparable with models trained with the citation form of the \textit{definiendum} that solely use context as a cue---viz. \citet{Gadetsky18WordDefGen} \& \citet{Chang18xSense}. 
All this makes $D_\text{Ctx}$ harder, but at the same time closer to a realistic application than the other two datasets, since each word appears inflected and in a specific \textsl{sentential context}. For applications of definition modeling, it would only be beneficial to take up these challenges; for example, the output ``\textit{monotremes: plural of monotreme}''\footnote{Definition from Wiktionary.} would not have been self-contained, necessitating a second query for ``\textit{monotreme}''.

\subsection{Results} \label{sec:results}

We use perplexity, a standard metric in definition modeling, to evaluate and compare our models.
Informally, perplexity assesses the model's confidence in producing the ground-truth output when presented the source input. It is formally defined as the exponentiation of cross-entropy.
We do not report \textsc{bleu} or \textsc{rouge} scores due to the fact that an important number of ground-truth definitions are comprised of a single word, in particular in $D_\text{Nor}$ ($\approx$ 25\%). Single word outputs can either be assessed as entirely correct or entirely wrong using \textsc{bleu} or \textsc{rouge}. However consider for instance the word ``\textit{elation}'': that it be defined either as ``\textit{mirth}'' or ``\textit{joy}'' should only influence our metric slightly, and not be discounted as a completely wrong prediction.

\begin{table}[ht]
    \centering
 \npdecimalsign{.}
 \nprounddigits{3}
    \begin{tabular}{l n{2}{3} n{2}{3} n{2}{3}}
         & {{$D_\text{Nor}$}} & {{$D_\text{Gad}$}} & {{$D_\text{Ctx}$}} \\
        \hline \hline
        \citeauthor{Noraset2017DefinitionML} & 48.168 &  45.62  & {{--}} \\
        \citeauthor{Gadetsky18WordDefGen} & {{--}} & 43.54 & {{--}} \\
        \hline
        Non-contextual & 42.1993 & 39.4279 & 48.2661 \\
        \textsc{Add} & {{--}} & 33.6775 & 43.695 \\
        \textsc{Select} & {{--}} & 33.9983 & 62.0387 \\
    \end{tabular}
    \caption{Results (perplexity)}
    \label{tab:res-ppl}
\end{table}

Table \ref{tab:res-ppl} describes our main results in terms of perplexity. 
We do not compare with \citet{Chang18xSense}, as they did not report the perplexity of their system and focused on a different dataset; likewise, \citet{Yang2019Sememes4CDM} consider only the Chinese variant of the task.
Perplexity measures for \citet{Noraset2017DefinitionML} and \citet{Gadetsky18WordDefGen} are taken from the authors' respective publications. 

All our models perform better than previous proposals, by a margin of 4 to 10 points, for a relative improvement of 11--23\%. Part of this improvement may be due to our use of Transformer-based architectures \citep{Vaswani17}, which is known to perform well on semantic tasks \citep[eg.]{Radford2018,Cer18USE,Devlin18Bert,Radford2019}.
Like \citet{Gadetsky18WordDefGen}, we conclude that disambiguating the \textit{definiendum}, when done correctly, improves performances: our best performing contextual model outranks the non-contextual variant by 5 to 6 points. The marking of the definiendum out of its context (\textsc{Add} vs.\ \textsc{Select}) also impacts results. Note also that we do not rely on task-specific external resources \citep[unlike][]{Noraset2017DefinitionML,Yang2019Sememes4CDM} or on pre-training \citep[unlike][]{Gadetsky18WordDefGen}.

Our 
contextual systems trained on the $D_\text{Gad}$ dataset used the concatenation of the \textit{definiendum} and the contextual cue as inputs. The definiendum was always at the start of the training example. This regular structure has shown to be useful for the models' performance: all models perform significantly worse on the more realistic data of $D_\text{Ctx}$ than on $D_\text{Gad}$. The $D_\text{Ctx}$ dataset is intrinsically harder for other reasons as well: it requires some form of lemmatization in every three out of eight training examples, and contains less data than other datasets, only half as many examples as $D_\text{Nor}$, and 20\% less than $D_\text{Gad}$.

The surprisingly poor results of \textsc{Select} on the $D_\text{Ctx}$ dataset may be partially blamed on the absence of a regular structure in $D_\text{Ctx}$. Unlike $D_\text{Gad}$, where the model must only learn to contextualize the first element of the sequence, in $D_\text{Ctx}$ the model has to single out the \textit{definiendum} which may appear anywhere in the sentence. Any information stored only in representations of contextual tokens will be lost to the decoders. The \textsc{Select} model therefore suffers of a bottleneck, which is highly regular in $D_\text{Gad}$ and that it may therefore learn to cope with; however predicting \textsl{where} in the input sequence the bottleneck will appear is far from trivial in the $D_\text{Ctx}$ dataset. We also attempted to retrain this model with various settings of hyperparameters, modifying dropout rate, number of warmup steps, and number of layers in the encoder---but to no avail. An alternative explanation may be that in the case of the $D_\text{Gad}$ dataset, the regular structure of the input entails that the first positional encoding is used as an additive marking device: only \textit{definienda} are marked with the positional encoding $\vec{\text{pos}(1)}$, and thus the architecture does not purely embrace a selective approach but a mixed one.

In any event, even on the $D_\text{Gad}$ dataset where the margin is very small, the perplexity of the additive marking approach \textsc{Add} is better than that of the \textsc{Select} model. This lends empirical support to our claim that definition modeling is a non-trivial sequence-to-sequence task, which can be better treated with sequence methods. The stability of the performance improvement  over the non-contextual variant in both contextual datasets also highlights that our proposed additive marking is fairly robust, and functions equally well when confronted to somewhat artificial inputs, as in $D_\text{Gad}$, or to linguistically coherent sequences, as in $D_\text{Ctx}$.

\section{Qualitative Analysis} \label{sec:quali}
\begin{table}[ht]
    \centering
 \subfloat[Handpicked sample]{
 \begin{tabular}{p{1.8cm} p{5cm}}
 \textbf{filch} & to seize \\
 \textbf{grammar} & the science of language \\
 \textbf{implosion} & a sudden and violent collapse \\
 \end{tabular}} \\
 \subfloat[Random sample]{
 \begin{tabular}{p{1.8cm} p{5cm}}
 \textbf{sediment} & to percolate \\
 \textbf{deputation} & the act of inciting \\
 \textbf{ancestry} & lineage \\
 \end{tabular}}
    \caption{Examples of production (non-contextual model trained on $D_\text{Nor}$)}
    \label{tab:generated-defs}
\end{table}


A manual analysis of definitions produced by our system reveals issues similar to those discussed by \citet{Noraset2017DefinitionML}, namely self-reference,\footnote{Self-referring definitions are those where a \textit{definiendum} is used as a \textit{definiens} for itself. Dictionaries are expected to be exempt of such definitions: as readers are assumed not to know the meaning of the \textit{definiendum} when looking it up.} \textsc{pos}-mismatches, over- and under-specificity, antonymy, and incoherence. Annotating distinct productions from the validation set, for the non-contextual model trained on $D_\text{Nor}$, we counted 9.9\% of self-references, 11.6\% \textsc{pos}-mismatches
, and 1.3\% of words defined as their antonyms. 
We counted \textsc{pos}-mismatches whenever the definition seemed to fit another part-of-speech than that of the \textit{definiendum}, regardless of both of their meanings; cf. Table\ \ref{tab:generated-defs} for examples.

\begin{table*}[t]
    \centering
    \begin{tabular}{p{2.75cm} p{6.5cm} p{4.75cm}}
        \textbf{Error type} & \textbf{Context} (\textit{definiendum} in bold) & \textbf{Production} \\
        \hline

\textsc{Pos}-mismatch &
her \textbf{major} is linguistics &
most important or important \\

Self-reference &
he wrote a letter of \textbf{apology} to the hostess &
a formal expression of apology \\

    \end{tabular}
    \caption{Examples of common errors (\textsc{Add} model trained on $D_\text{Nor}$)}
    \label{tab:errors}
\end{table*}

For comparison, we annotated the first 1000 productions of the validation set from our \textsc{Add} model trained on $D_\text{Ctx}$. We counted 18.4\% \textsc{pos} mismatches and 4.4\% of self-referring definitions; examples are shown in Table \ref{tab:errors}.
The higher rate of \textsc{pos}-mismatch may be due to the model's hardship in finding which word is to be defined since the model is not presented with the \textit{definiendum} alone: access to the full context may confuse it.
On the other hand, the lower number of self-referring definitions may also be linked to this richer, more varied input: this would allow the model not to fall back on simply reusing the \textit{definiendum} as its own \textit{definiens}.
Self-referring definitions highlight that our models equate the meaning of the \textit{definiendum} to the composed meaning of its \textit{definientia}. Simply masking the corresponding output embedding might suffice to prevent this specific problem; preliminary experiments in that direction suggest that this may also help decrease perplexity further. 

As for \textsc{pos}-mismatches, we do note that the work of \citet{Noraset2017DefinitionML} had a much lower rate of 4.29\%: we suggest that this may be due to the fact that they employ a learned character-level convolutional network, which arguably would be able to capture orthography and rudiments of morphology. Adding such a sub-module to our proposed architecture might diminish the number of mistagged \textit{definienda}. Another possibility would be to pre-train the model, as was done by \citet{Gadetsky18WordDefGen}: in our case in particular, the encoder could be trained for \textsc{pos}-tagging or lemmatization.

Lastly, one important kind of mistakes we observed is hallucinations. Consider for instance this production by the \textsc{Add} model trained on $D_\text{Ctx}$, for the word ``\textit{beta}'': ``\textit{the twentieth letter of the Greek alphabet ($\kappa$), transliterated as `o'.}''. Nearly everything it contains is factually wrong, though the general semantics are close enough to deceive an unaware reader.\footnote{On a related note, other examples were found to contain unwanted social biases; consider the production by the same model for the word ``\textit{blackface}'': ``\textit{relating to or characteristic of the theatre}''. Part of the social bias here may be blamed on the under-specific description that omits the offensive nature of the word; however contrast the definition of Merriam Webster for \textit{blackface}, which includes a note on the offensiveness of the term, with that of Wiktionary, which does not. Cf. \citet{Bolukbasi2016,swinger2018biases} for a discussion on biases within embedding themselves.} 
We conjecture that filtering out hallucinatory productions will be a main challenge for future definition modeling architectures, for two main reasons: firstly, the tools and metrics necessary to assess and handle such hallucinations have yet to be developed; secondly, the input given to the system being word embeddings, research will be faced with the problem of grounding these distributional representations---how can we ensure that ``\textit{beta}'' is correctly defined as ``\textit{the second letter of the Greek alphabet, transliterated as `b'}'', if we only have access to a representation derived from its contexts of usage? Integration of word embeddings with structured knowledge bases might be needed for accurate treatment of such cases.

\section{Conclusion} \label{sec:ccl}

We introduced an approach to generating word definitions that allows the model to access rich contextual information about the word token to be defined. Building on the distributional hypothesis, we naturally treat definition generation as a sequence-to-sequence task of mapping the word's context of usage (input sequence) into the context-appropriate definition (output sequence).


We showed that our approach is competitive against a more naive `contextualize and select' pipeline. This was demonstrated by comparison both to the previous contextualized model by \citet{Gadetsky18WordDefGen} and to the Transformer-based \textsc{Select} variation of our model, which differs from the proposed architecture only in the context encoding pipeline. While our results are encouraging, given the existing benchmarks we were limited to perplexity measurements in our quantitative evaluation. A more nuanced semantically driven methodology might be useful in the future to better assess the merits of our system in comparison to alternatives. 


Our model opens several avenues of future explorations. One could  straightforwardly extend it to generate definitions of multiword expressions or phrases, or to analyze vector compositionality models by generating paraphrases for vector representations produced by these algorithms. Another strength of our approach is that it can provide the basis for a standardized benchmark for contextualized and non-contextual embeddings alike: downstream evaluation tasks for embeddings systems in general either apply to non-contextual embeddings \citep[eg.]{Gladkova2016} or to contextual embeddings \citep[eg.]{wang2019glue} exclusively, redefining definition modeling as a sequence-to-sequence task will allow in future works to compare models using contextual and non-contextual embeddings in a unified fashion. Lastly, we also intend to experiment on languages other than English, especially considering that the required resources for our model only amount to a set of pre-trained embeddings and a dataset of definitions, either of which are generally simple to obtain. 

While there is a potential for local improvements, 
our approach has demonstrated its ability to account for contextualized word meaning in a principled way, while training contextualized token encoding and definition generation end-to-end.
%
Our implementation is efficient and fast, building on free open source libraries for deep learning, and shows good empirical results. Our code, trained models, and data will be made available to the community.

\section*{Acknowledgments}
We thank Quentin Gliosca for his many remarks throughout all stages of this project. We also thanks Kees van Deemter, as well as anonymous reviewers, for their thoughtful criticism of this work. 
The work was supported  by a public grant overseen by the French National Research Agency (ANR)  as part of the ``Investissements d'Avenir'' program: Idex \emph{Lorraine Universit\'e d'Excellence} (reference: ANR-15-IDEX-0004).

\bibliography{tmickus-refs}
\bibliographystyle{acl_natbib}

\end{document}